\documentclass[conference]{IEEEtran}
\usepackage[T1]{fontenc}
\usepackage{graphicx}
\usepackage{booktabs}
\usepackage{amsmath}
\usepackage{xcolor}
\usepackage[hidelinks]{hyperref}
\usepackage{url}
\hypersetup{pdftitle={Full-Weight VLA Fine-Tuning Without an fp32 Master Copy Silently Erases Most Updates}, pdfauthor={Shubham Singh},
  pdfkeywords={bfloat16, mixed precision, master weights, AdamW, vision-language-action models, LeRobot}}

\newcommand{\lr}{\eta}
\begin{document}

\title{Full-Weight VLA Fine-Tuning Without an fp32 Master Copy Silently Erases Most Updates}

\author{\IEEEauthorblockN{Shubham Singh}
\IEEEauthorblockA{singh281@utexas.edu}}

\maketitle

\begin{abstract}
When a vision-language-action (VLA) model is fine-tuned with its trainable weights in bfloat16 (bf16) and no fp32
master copy, AdamW writes each update straight into bf16, silently rounding away any update smaller than 1/512 of a
weight. In 9 of 12 open VLA codebases a default or documented recipe trains this way, 4 on full weight matrices,
including LeRobot's SmolVLA default and its $\pi_{0}$/$\pi_{0.5}$ guides. In our LIBERO runs without the copy, a probe
every 100 steps finds at least 79\% of updates to bf16 weights erased. On the Hugging Face Hub, 99\% of sampled
full-weight SmolVLA fine-tunes store the action expert in bf16, and 84\% of recent $\pi_{0.5}$ fine-tunes with a
training configuration set the bf16 flag. On LIBERO-Object, with rules fixed in advance, an fp32 copy raises success
over LeRobot's storage from 4.3\% to 49.3\% for SmolVLA and from 5.1\% to 80.9\% for $\pi_{0.5}$ at a ten-times lower
learning rate, where both models are undertrained, and from 89.0\% to 97.7\% for $\pi_{0.5}$ at its default rate. For
SmolVLA at its default rate it adds 10.2 points with open-loop 50-action chunks but only 4.2 when re-planning every 10
actions. In these five comparisons and two more against SmolVLA with every trainable weight in bf16, every run with the
copy beats every run without it ($p<0.05$); all gaps but the 8.7- and 4.2-point ones clear our 10-point threshold. The
fix is to keep the trainable weights in fp32 during training, a one-line change for SmolVLA in LeRobot.
\end{abstract}

\vspace{3pt}\noindent\fbox{\parbox{\dimexpr\columnwidth-2\fboxsep-2\fboxrule\relax}{\footnotesize\raggedright
\textbf{Am I affected?} Yes, if you fine-tune full weight matrices that are bf16 when the optimizer is built
(\texttt{\{p.dtype for p in model.parameters() if p.requires\_grad\}} shows \texttt{torch.bfloat16}) and the optimizer,
like PyTorch's AdamW, keeps no fp32 copy, as in LeRobot's SmolVLA default and the \texttt{-{}-policy.dtype=bfloat16} of
its $\pi_{0}$/$\pi_{0.5}$ guides, openpi's PyTorch trainer and RLinf (Table~\ref{tab:audit}). LoRA adapters stored in
bf16 keep updates of learning-rate size at their default rates of $2$--$5{\times}10^{-4}$; we did not measure smaller
ones.\\[3pt]
\textbf{Fix:} keep those weights in fp32 and compute in bf16 under autocast. In LeRobot, for $\pi_{0}$/$\pi_{0.5}$,
set \texttt{-{}-policy.dtype=float32} and add \texttt{-{}-accelerator.mixed\_precision=bf16}; for SmolVLA, which by
default trains only its action expert, pass \texttt{dtype=torch.float32} where \texttt{smolvlm\_with\_expert.py} builds
the expert, and add the same flag. Memory and learning rate: Section~\ref{sec:limits}.}}

\section{Introduction}

Fine-tuning a pretrained VLA on a new task is now routine, done with a few open codebases: LeRobot
\cite{lerobot,a2602.22818} for SmolVLA \cite{a2506.01844} and $\pi_{0}$/$\pi_{0.5}$ \cite{a2410.24164,a2504.16054},
openpi \cite{openpi}, the OpenVLA family \cite{a2406.09246,a2502.19645,a2509.09372}, and RL frameworks such as RLinf
\cite{a2509.15965}. To save memory, several cast the trainable weights to bfloat16 (bf16) before building the optimizer.
Compute in bf16 is intended, but PyTorch's AdamW then also keeps its moments in bf16 and adds each update directly to
the bf16 weight, which mixed-precision training avoids by keeping an fp32 \emph{master copy} of the weights and rounding
only the copy used for compute \cite{a1710.03740}. Without the copy, an update smaller than 0.2--0.4\% of the weight is
rounded away \cite{a1502.02551}. Fine-tuning learning rates are $10^{-5}$ to $10^{-4}$ and most AdamW steps are smaller
still, so for weights of typical size a large share of updates is silently lost, with no error and a falling loss.

We first state when a statistic $E$, computed from the weights alone \cite{a2602.03839,dreamzero31}, is a lower bound on
the share of weights an AdamW step leaves unchanged, and check it with a per-step probe (Section~\ref{sec:rounding}). We
then audit 12 open VLA codebases and sample fine-tunes on the Hugging Face Hub (Section~\ref{sec:audit}). Finally, on
LIBERO-Object \cite{a2306.03310}, with rules fixed in advance, we compare three ways to store the trainable weights
(Section~\ref{sec:success}). Two keep no fp32 copy: \emph{all-bf16} stores them all in bf16, and \emph{LeRobot's
recipe} stores most of them in bf16 (SmolVLA's default, or the bf16 flag of LeRobot's $\pi_{0.5}$ guide). The third,
\emph{the fp32 copy}, stores them all in fp32.

\section{How an update is rounded away}\label{sec:rounding}

A bf16 number keeps 8 significant bits, so the bf16 numbers next to a weight $w$ are $|w|/256$ to $|w|/128$ apart.
Rounding to the nearest gives back $w$ for any update smaller than half that gap: every update smaller than $|w|/512$
is lost, and updates up to $|w|/256$ are lost for some weights \cite{a1502.02551,a2010.06192}. For example, an update of
$10^{-5}$ cannot change any weight with $|w|>2^{-8}\approx0.004$, which covers 88\% of SmolVLA's action-expert weights.

\begin{figure}[t]
\centering
\includegraphics[width=\columnwidth]{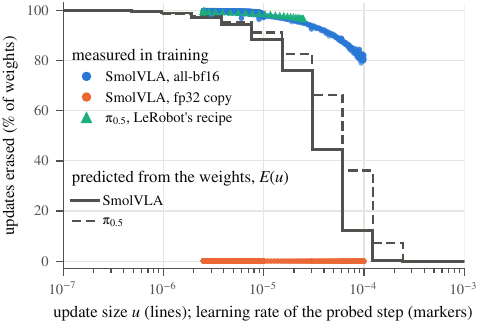}
\caption{\textbf{Without an fp32 copy, measured erasure always exceeds the share predicted from the weights.} Lines:
$E(u)$ from released weights (solid: SmolVLA's action expert in \texttt{smolvla\_base}; dashed: the weight matrices that
$\pi_{0.5}$'s documented recipe stores in bf16, in \texttt{pi05\_libero\_finetuned\_v044}). Markers: the share of those
weights unchanged after an AdamW step, every 100 steps, at that step's learning rate; for $\pi_{0.5}$, the bf16 weights
in tensors that received a gradient, including the token embedding, which the dashed line leaves out. SmolVLA's runs of
LeRobot's recipe, also above $E$, are left out for legibility.}
\label{fig:erasure}
\end{figure}

\textbf{Erased share from the weights.} $E(u)$ is the share of weights that an update of size $u$ cannot change,
whichever its sign: both $w+u$ and $w-u$ round back to $w$ in bf16. $E$ needs only the weights (minutes on a CPU) and
changes only where $u$ crosses a power of two (the steps in Fig.~\ref{fig:erasure}). Such estimates have been made
before \cite{a2602.03839,dreamzero31,rlinf1649}; here we state when $E$ is a bound.

\textbf{When $E$ is a lower bound.} Rounding is monotone, so if no element of a step moves by more than $c\lr_t$, where
$\lr_t$ is the step's learning rate, at least a share $E(c\lr_t)$ of the weights stays unchanged. AdamW moves an element
by $\lr_t\,\hat m/(\sqrt{\hat v}+\epsilon)$ plus decoupled weight decay \cite{a1711.05101}. If $\beta_1^2<\beta_2$, the
Cauchy--Schwarz inequality bounds $|\hat m|/\sqrt{\hat v}$ by $c=(1-\beta_1)/\sqrt{(1-\beta_2)(1-\beta_1^2/\beta_2)}$
at every step \cite{a2602.03839}: 1.165 for the $\beta=(0.9,0.95)$ of LeRobot, openpi and RLinf's $\pi_0$ recipe, but
7.3 for PyTorch's default $\beta_2=0.999$, for which $E$ is only an estimate (marked in Table~\ref{tab:audit}). With the
moments stored in bf16, a search over gradients between 1/4 and 8 times the first moment found steps up to $1.202\lr$,
never $1.21\lr$; for every peak learning rate here the next power of two is at least 1.22 times larger, so
$E(1.21\lr)=E(\lr)$, and since $\lr_t\le\lr$ this bounds every step, as far as that search reaches. In LeRobot and
openpi, weight decay is a separate factor $1-\lr\lambda$, which moves $w$ by less than $|w|/512$ and so is itself rounded
away (RLinf's fused AdamW adds the decay to the step before rounding; at its $5\times10^{-6}$ the bound still holds).
Strictly, the bound holds for $E$ of the current weights; we compute $E$ once, from released weights, and the probe
checks it throughout.

\textbf{Measured erased share.} A probe copies the trainable weights before every 100th optimizer step and counts the
elements whose stored value is unchanged afterwards (Fig.~\ref{fig:erasure}). Before any run we fixed a criterion: the
measured share must be at least $E(\lr_t)-2$ points at every probe after the first 20 ($E$ then averaged over the two
signs). It holds in all 6 all-bf16 runs and in the 6 runs of LeRobot's recipe for SmolVLA; the all-bf16 runs in fact
exceed $E(\lr_t)$ at every probe by at least 1.2 points. Most AdamW steps are much smaller than $\lr_t$, so at
$\lr=10^{-4}$, where $E=12\%$, 79--99\% of the updates are erased, and 96--99.95\% at $10^{-5}$. With SmolVLA's fp32
copy at most 0.2\% of elements stay unchanged in a step. For $\pi_{0.5}$ the probe counts the bf16 weights in tensors
that receive a gradient: at every probe 96.7--99.7\% of their updates are erased (at least 95.8\% without the token
embedding, most of whose rows get no gradient), against $E(2.5\times10^{-5})=83\%$ from
\texttt{pi05\_libero\_finetuned\_v044} ($E$ is about 1.5 points lower on three sampled weight matrices of the
\texttt{pi05\_base} weights our runs start from); at $2.5\times10^{-6}$, 99.4--99.8\% (at least 99.2\% without the token
embedding) against $E=97.7\%$, so the criterion, fixed before those runs, holds at every probe.

\begin{table}[t]
\caption{Default VLA fine-tuning recipes that store trainable weights in bf16 with no fp32 copy. $E$: share of weights
that an update the size of the learning rate cannot change; a lower bound on the share left unchanged per step
(Section~\ref{sec:rounding}).}
\label{tab:audit}
\centering
\setlength{\tabcolsep}{3pt}
\footnotesize
\begin{tabular}{@{}llrr@{}}
\toprule
Codebase: recipe & Model (size) & Default $\lr$ & $E(\lr)$ \\
\midrule
LeRobot: default$^\dagger$ & SmolVLA (0.45B) & $10^{-4}$ & 12\% \\
LeRobot: as documented$^\dagger$ & $\pi_{0.5}$ (4.1B) & $2.5{\times}10^{-5}$ & 83\% \\
openpi: PyTorch$^\dagger$ & $\pi_{0}$ & $2.5{\times}10^{-5}$ & 78\% \\
openpi: PyTorch$^\dagger$ & $\pi_{0.5}$ & $5{\times}10^{-5}$ & 65\% \\
RLinf: RL (PPO) & $\pi_{0}$ & $5{\times}10^{-6}$ & 94\% \\
RLinf: RL & OpenVLA-OFT (7.5B) & $2{\times}10^{-5}$ & 71\%$^\ast$ \\
Pion: full, AdamW arm & VLA-Adapter (0.5B) & $10^{-4}$ & 15\%$^\ast$ \\
5 codebases$^\ddagger$: LoRA & 0.5--7B & $2$--$5{\times}10^{-4}$ & ${\approx}0$\%$^{\ast\S}$ \\
\bottomrule
\end{tabular}\\[3pt]
\parbox{\columnwidth}{\footnotesize $^\dagger$Confirmed by building the training setup; the other rows by code reading.
$^\ast$An estimate, not a bound: AdamW with $\beta_2=0.999$. $^\ddagger$OpenVLA, OpenVLA-OFT, MiniVLA, VLA-Adapter and
UniVLA, which share OpenVLA's LoRA code and PEFT version. $^\S$Updates of the learning rate's size survive; we did not
measure smaller ones. $E$ is from released checkpoints (LeRobot's $\pi_{0.5}$ row:
\texttt{pi05\_libero\_finetuned\_v044}; LoRA: adapters at initialization); not computed for LeRobot's $\pi_{0}$ guide,
which uses the same flag. Not affected (fp32 weights or a master copy): GR00T (NVIDIA's trainer and LeRobot's$^\dagger$),
SimpleVLA-RL, CogACT, openpi's JAX trainer, OpenVLA's full fine-tuning, and LeRobot's $\pi_{0}$/$\pi_{0.5}$ config
default$^\dagger$.}
\end{table}

\section{Audit of VLA fine-tuning codebases}\label{sec:audit}

We read the default (or documented) fine-tuning entry point of 12 open VLA codebases at pinned commits (LeRobot
\cite{lerobot}, openpi \cite{openpi}, RLinf \cite{a2509.15965}, Pion \cite{a2605.19282}, OpenVLA \cite{a2406.09246},
OpenVLA-OFT \cite{a2502.19645}, MiniVLA \cite{minivla}, VLA-Adapter \cite{a2509.09372}, UniVLA \cite{a2505.06111}, Isaac
GR00T \cite{a2503.14734}, SimpleVLA-RL \cite{a2509.09674} and CogACT \cite{a2411.19650}), recorded the dtype in which the
trainable weights are \emph{stored} when the optimizer is built and whether an fp32 copy exists, and confirmed the
LeRobot and openpi rows at runtime by building the training setup. In 9 of the 12, a default or documented recipe stores
trainable weights in bf16 with no copy (Table~\ref{tab:audit}). Four (LeRobot, openpi's PyTorch trainer, RLinf and Pion)
train full weight matrices and lose 12--94\% of learning-rate-sized updates; RLinf at least logs a warning that
recommends fp32. Five OpenVLA-derived scripts train LoRA adapters (two also a bf16 action head) at rates where updates of
that size survive ($E\approx0$). We did not measure smaller steps, such as those of OpenVLA-OFT's last 50{,}000 steps at
a tenfold lower rate, where $E$ is 28\% for its released LoRA weights (an estimate).

Two lessons came from the runtime checks of LeRobot. First, code reading had marked the SmolVLA default safe, but at
runtime it keeps 96.6M of its 99.9M trainable weights in bf16 with no copy: the action expert is built from the
vision-language model's config, which carries bf16, and SmolVLA accepts \texttt{-{}-policy.dtype=float32} but ignores
it. Only 3.3M weights, in layers SmolVLA builds outside that config, stay in fp32. Second, LeRobot's
$\pi_{0}$/$\pi_{0.5}$ policies default to fp32, but the fine-tuning commands in its $\pi_{0}$ and $\pi_{0.5}$ guides add
\texttt{-{}-policy.dtype=bfloat16} and call it mixed precision; that stores 3.61B trainable weights in bf16 with no copy.
LeRobot's general accelerator page documents fp32 weights with bf16 compute, but these guides do not use it. openpi's
README states that its PyTorch trainer trains in full bf16 by default and reaches higher losses than in fp32.

\textbf{Fine-tunes on the Hugging Face Hub.} On 16 October 2026 the Hub listed 8{,}196 public models tagged as
fine-tunes of \texttt{lerobot/smolvla\_base} and 825 tagged as fine-tunes of \texttt{lerobot/pi05\_base}. With a method
and checks fixed in advance (after a smaller pilot), we read the file headers of random samples, and six weight matrices
where needed, without downloading the models. Of 300 SmolVLA fine-tunes, 291 store full weights with SmolVLA's action
expert, and 289 of these (99.3\%) store the expert's weight matrices in bf16, as LeRobot's default trains them. In 40 of
the bf16 fine-tunes whose training configuration starts from \texttt{smolvla\_base}, all at its default learning rate of
$10^{-4}$, a median 22\% of the elements of six expert weight matrices are still bit-identical to the base after
training; all 40 exceed the 11.6\% that no step at this learning rate can change (a check we added afterwards), and a
model trained from scratch gives 0.1\%. Of 300 randomly drawn \texttt{pi05\_base} fine-tunes from the last eight weeks
that include a training configuration, 252 (84\%) set LeRobot's bf16 flag, 242 of them on full weights, and the 10 of
those 242 we checked store their language-model and action-expert weights in bf16. Base-model tags are chosen by the
uploaders; a second reading with independent code agreed on all 20 SmolVLA repositories it re-measured.

\begin{figure}[t]
\centering
\includegraphics[width=\columnwidth]{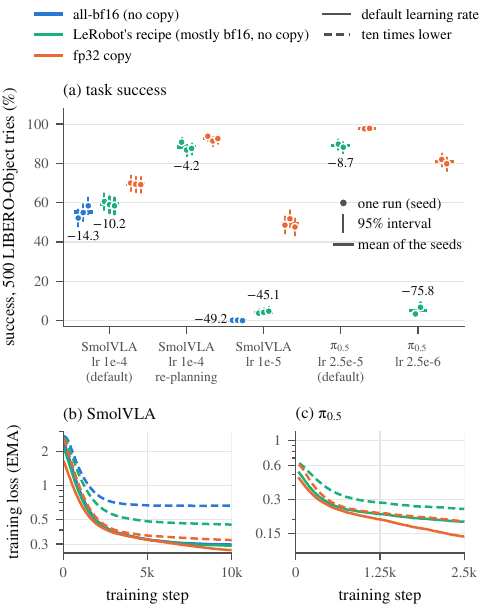}
\caption{\textbf{The fp32 copy raises LIBERO-Object success in every comparison, most at the ten-times lower learning
rate.} Colors and line styles as in the legend (all-bf16: SmolVLA only). (a) Success over 500 tries per run: dots are
seeds with Wilson 95\% intervals (scoring noise only), bars the means of the seeds, and numbers each blue or green
cluster's mean minus the orange mean of its group, in points; SmolVLA executes 50-action chunks open loop except in the
second group (retrained, re-planning every 10 actions). (b, c) Training loss, mean over seeds (exponential moving
average).}
\label{fig:results}
\end{figure}

\section{Effect on task success}\label{sec:success}

\textbf{Setup.} We fine-tune SmolVLA \cite{a2506.01844} from \texttt{smolvla\_base} with LeRobot (commit
\texttt{b9cb121}) on the 454 LIBERO-Object demonstrations for 10{,}000 steps at batch 16 (LeRobot's SmolVLA guide uses
20{,}000 at batch 64; its LIBERO page and the SmolVLA paper, 100{,}000 at batch 64 on all four LIBERO suites), with
AdamW ($\beta=(0.9,0.95)$), LeRobot's warmup and cosine schedule, bf16 autocast, and learning rates $10^{-4}$ (the
default) and $10^{-5}$. Only the storage of the 99.9M trainable weights differs: all-bf16, LeRobot's recipe (its
default: 3.3M in fp32, 96.6M in bf16) or the fp32 copy. AdamW keeps the moments in the weight's dtype, but with
$\beta_2=0.95$ bf16 moments change AdamW's mean step by less than 0.2\% in a simulation. All settings are scored with
the same inference dtypes on 500 tries (10 tasks, 50 fixed initial states each), executing each predicted chunk of 50
actions open loop (SmolVLA's default; the SmolVLA paper re-plans after every action and reports 96\%). Before training
we fixed the rule: a setting effect is large (``real'' in our plan) if the mean success of two settings differs by at
least 10 points \emph{and} by more than the larger seed-to-seed gap within a setting. All-bf16 and the fp32 copy ran
first, with seeds 1000 and 1001; LeRobot's recipe and a third seed, 1002, for all three settings came later, with their
analysis fixed before running them. Each rule and prediction, and every amendment but one (to the scorer check), went
into a dated project log before the runs it covers; the log was not filed with an outside registry. The paired tests
were fixed in advance, except for the first eight SmolVLA runs and $\pi_{0.5}$ at its default rate, where we added them
after scoring.

\textbf{Scorer checks.} Our scorer passes its check for $\pi_{0.5}$ but not for SmolVLA, so SmolVLA's success rates
compare settings only within this paper. On an H100 it gives 99.2\% for LeRobot's published
\href{https://huggingface.co/lerobot/pi05_libero_finetuned_v044}{\texttt{lerobot/\allowbreak pi05\_libero\_\allowbreak finetuned\_v044}},
reported at 99.0\%. The published
\href{https://huggingface.co/lerobot/smolvla_libero}{\texttt{lerobot/smolvla\_libero}} scores 67.6\% with its 50-action
chunks, against the 88\% we required in advance, but 93.4\% when it re-plans every 10 actions, so its shortfall comes
from its chunks, not the simulator.
\href{https://huggingface.co/HuggingFaceVLA/smolvla_libero}{\texttt{HuggingFaceVLA/smolvla\_libero}}, which re-plans
after every action and which we added after the first five tasks of that check fell short, scored 81.4\% (exact 95\%
interval 76.3--85.8\%) on five and a half tasks before we stopped it; its shortfall is unexplained, and the scorer
debugging our plan called for after a second miss was not done.

\textbf{Results.} Scored with 50-action chunks, every SmolVLA comparison passes the rule (Fig.~\ref{fig:results}a).
Against LeRobot's recipe, the fp32 copy adds 10.2 points at $10^{-4}$ (69.5 against 59.3\%, seed spread 2.2), just
above the threshold, and 45.1 points at $10^{-5}$ (49.3 against 4.3\%, spread 4.2; gaps are differences of unrounded
means). Against all-bf16 it adds 14.3 points (from 55.2\%, spread 6.2) and 49.2 points (from 0.1\%, spread 4.2), or
16.1 and 50.0 with the first two seeds alone, as first planned. Before the first runs we had predicted a small or no gap
at $10^{-4}$, where $E$ is only 12\%; the probe's 79--99\% erasure suggests why it is larger. Here and below, every run
with the copy beats every run without it, and the copy scores higher on all 10 tasks, summed over the seeds. All
settings face the same 500 initial states and environment seeds, so the tries are paired: pooled over the seeds, the
fp32 copy succeeds on 370 tries where LeRobot's recipe fails, against 217 the other way, at $10^{-4}$ (exact McNemar
$p=3\times10^{-10}$), and on 701 against 25 at $10^{-5}$, and each of the four gaps is significant at each seed
($p<0.002$). These tests compare trained models, but counting seed-to-seed variation changes little: across all 26
runs, runs of the same setting and learning rate scatter about as much as 500 independent tries would (dispersion 0.97,
95\% interval 0.54--2.25), and a quasi-binomial model of the run totals, added after all runs were scored, gives
$p<3\times10^{-5}$ for the six gaps without re-planning in Fig.~\ref{fig:results}a ($p<0.002$ at the interval's upper
end; $p=0.007$ for $\pi_{0.5}$'s default-rate gap with its own, larger run-to-run variation). Retraining all-bf16's
seed-1000 run at $10^{-4}$ on an RTX 5060 instead of an RTX 4090 gave 48.2\% against 52.2\%: the GPU alone moves a run
about as much as a seed does. LeRobot's 3.3M fp32 weights recover 4.1 points over all-bf16 at both rates, 8\% of the gap
at $10^{-5}$, but at $10^{-4}$ less than all-bf16's seed spread, so we cannot tell whether they help there. At $10^{-5}$
the all-bf16 loss falls only from 0.69 to 0.66 after step 2{,}000, while the fp32 copy reaches 0.32 and LeRobot's
recipe 0.45 (Fig.~\ref{fig:results}b; means over 250 steps), yet LeRobot's recipe succeeds on 4.3\% of tries; even the
fp32 copy ends 20.2 points below its default-rate score, so the gaps at this rate are measured on undertrained models.
At $10^{-4}$ the final losses are close (0.30 and 0.29 for the two bf16 settings against 0.26), yet success differs by
10--14 points: the loss curve does not show the lost success.

\textbf{SmolVLA with re-planning.} Open-loop 50-action chunks are SmolVLA's weakest scoring (above), and our trained
models were not kept, so we retrained the six default-rate runs of LeRobot's recipe and the fp32 copy (same seeds,
scripts and GPU type) and scored each model both ways on the same 500 tries, with the rule and what each outcome would
change in this paper fixed in advance. With 50-action chunks the retrained runs reproduce the earlier ones: in each run
498--500 of the 500 tries end the same way. Re-planning every 10 actions shrinks the gap to 4.2 points (88.4 against
92.6\%, seed spread 4.0), below the threshold, though the copy still scores higher at each seed and on all 10 tasks, and
succeeds on 129 tries where LeRobot's recipe fails, against 66 the other way (exact McNemar $p=8\times10^{-6}$, fixed
in advance). With seed-to-seed variation counted as above (added after scoring), $p=0.001$, and 0.04 with these six
runs' own dispersion. Our recorded prediction had favored a gap below 3 points (probability 0.45).

\textbf{$\pi_{0.5}$ with LeRobot's documented bf16 flag.} We repeat the comparison on $\pi_{0.5}$ \cite{a2504.16054}
with the flags of LeRobot's $\pi_{0.5}$ LIBERO quickstart (which re-plans every 10 actions) and four changes: the
general \texttt{lerobot/pi05\_base} instead of the quickstart's LIBERO-trained base, which already solves
LIBERO-Object; the 454 LIBERO-Object demonstrations instead of all of LIBERO; 2{,}500 steps at batch 32 instead of
30{,}000 at batch 64; and bf16 autocast. The flag stores 3.61B of the 4.14B trainable weights in bf16 with no copy
(LeRobot's recipe); the fp32 copy loads with the same flag and casts them back to fp32, so both settings start from the
same weights, with those 3.61B rounded to bf16. Both use LeRobot's default learning rate $2.5\times10^{-5}$ with its
warmup and cosine schedule, seeds 1000 and 1001, and AdamW updating one tensor at a time. LeRobot's published
checkpoint above was itself fine-tuned with the bf16 flag, but for 6{,}000 steps from Physical Intelligence's LIBERO
model, which openpi reports at 98.2\% on LIBERO-Object \cite{openpi}, so it had little left to learn.

The fp32 copy raises success from 89.0 to 97.7\% (Fig.~\ref{fig:results}a): $+8.7$ points, seed spread 1.6, of the at
most 11 points LeRobot's recipe leaves to gain, so the copy removes 79\% of its failures. It is higher on all 10 tasks
and succeeds on 108 tries where LeRobot's recipe fails, against 21 the other way (exact McNemar $p<10^{-12}$;
$p<10^{-6}$ at each seed). The gain is below the 10-point threshold, so by our rule it does not count as a large effect.
Both borderline verdicts could have gone either way: the 95\% intervals of the gaps over the paired tries, 6.6--10.9
points here and 7.1--13.3 for SmolVLA's 10.2, both contain 10. The final losses are 0.19 against 0.14
(Fig.~\ref{fig:results}c).

\textbf{$\pi_{0.5}$ at a ten-times lower learning rate.} Because the default-rate gap fell short of the threshold, we
added a second $\pi_{0.5}$ comparison at $2.5\times10^{-6}$, a rate our first plan included for every model but we had
dropped for $\pi_{0.5}$ to save money. Before running it we fixed the same rule and analysis and predicted a gap of at
least 10 points. Only the learning rate and the GPU change; LeRobot's schedule ends at $2.5\times10^{-6}$, so the rate
stays there after warmup, as for any LeRobot user who lowers it. LeRobot's recipe scores 5.1\% and the fp32 copy
80.9\% (Fig.~\ref{fig:results}a): $+75.8$ points with a seed spread of 3.4, a large effect by our rule. The copy
succeeds on 765 tries where LeRobot's recipe fails, against 7 the other way (exact McNemar $p<10^{-12}$, also at each
seed), and is higher on all 10 tasks. As for SmolVLA at $10^{-5}$, the loss gives no warning: LeRobot's recipe still
lowers it, to 0.25 against 0.19 (Fig.~\ref{fig:results}c). At this rate 2{,}500 steps leave even the fp32 runs 17
points below their default-rate score (scored on another GPU type), so this gap is measured on undertrained models.

\textbf{Compute.} Rented GPUs: one RTX 4090 per SmolVLA run; one H100 SXM per $\pi_{0.5}$ run at the default learning
rate and, because most rented H100s failed our scoring-speed check, one RTX PRO 6000 Blackwell at the lower rate, on
which we did not re-score the published checkpoint. The 26 runs used about 19 RTX 4090-hours, 8.4 H100-hours and 13 RTX
PRO 6000-hours, including setup and scoring; the six retrained for re-planning about 6 more RTX 4090-hours; and scorer
checks, development and pods that failed our setup checks about 6 more RTX 4090-hours, 4 more H100-hours and under one
A40-hour (about \$110 of rented compute in all). Local development used one RTX 5060 (8\,GB).

\section{Related work}

Mixed-precision training keeps fp32 master weights because fp16 weights alone lost small updates and accuracy
\cite{a1710.03740}; with the copy, bf16 matches fp32 \cite{a1905.12322}. Without it, training can be rescued by
stochastic rounding or Kahan summation \cite{a1502.02551,a2010.06192,a2102.13565}, by compensating the rounding error
\cite{a2601.22101,a2405.03637}, which otherwise grows as the learning rate shrinks \cite{a2601.22101}, or by
multiplicative updates \cite{a2607.10611}, and stochastic write-back recovers most of the loss gap \cite{a2607.09800};
in LLM continued pretraining, pure bf16 has been found adequate \cite{a2408.15793}, but in distilling a 4B-parameter
language model with AdamW at $\beta=(0.9,0.95)$ and a learning rate of $10^{-5}$, an fp32 master copy raised micro-F1
from 0.143 to 0.187 over bf16 weights without one (three seeds) \cite{a2606.06840}. In RL post-training of language
models with an fp32 master copy, about 99\% of per-step updates are invisible after the bf16 cast used for compute,
which its authors trace to Adam's steps falling below the bf16 rounding threshold \cite{a2602.03839}; lost or invisible
bf16 updates are also reported for on-policy distillation \cite{a2610.02179}, GRPO \cite{a2610.08430}, full
fine-tuning of small language models, where 77--98\% of weights stored in bf16 never changed \cite{a2610.22438}, and
ternary language models, at checkpoint export and during fine-tuning \cite{a2610.07853}. For robot policies, Silent Collapse \cite{silentcollapse} finds silent
failures under bf16 autocast that arise in the gradient chain; it does not study weights stored in bf16. A pull request
porting DreamZero to a LeRobot fork estimates from the released weights that bf16 weights would lose about 84\% of
updates at $10^{-5}$, without measuring success \cite{dreamzero31}. GuidedVLA reports that full-bf16 training of
$\pi_{0}$ scores 10 against 90 with fp32 or mixed precision \cite{a2605.12369}, and an OpenTau issue reports
$\pi_{0.5}$ on LIBERO at 0\% without and 94\% with fp32 master weights \cite{opentau181}. In RL fine-tuning, most
action-expert weights of SmolVLA \cite{a2609.28838} and $\pi_{0}$ \cite{rlinf1649} stay bit-identical across updates in
bf16; the SmolVLA study also lists the precision settings of RLinf, openpi and LeRobot's SmolVLA \cite{a2609.28838}. An
RLinf pull request added a $\pi_{0.5}$ SFT path with fp32 master weights so that warmup updates are not lost
\cite{rlinf1254}. These VLA results are single runs, at one learning rate, or in RL; ours are seeded supervised
comparisons at two learning rates, with rules fixed in advance, on two VLAs including LeRobot's own recipes.

\section{Limitations and recommendations}\label{sec:limits}

The evidence is narrow: two models, one LIBERO suite, three seeds per SmolVLA setting and two per $\pi_{0.5}$ setting,
shortened schedules for both models ($\pi_{0.5}$'s from the general base), and SmolVLA scored mostly on open-loop
50-action chunks. The probe, the stall at $10^{-5}$ and the larger gaps at the lower learning rate point to the erased
updates rather than the bf16 moments that come with them, though we did not ablate the two. How large the gain is for
other VLAs, tasks or schedules is untested. We did not re-plan at $10^{-5}$ or for all-bf16, train with the guide's
longer schedule, or run LeRobot's $\pi_{0.5}$ recipe from its LIBERO-trained base, where its published bf16 checkpoint
already scores 99.2\%. Our runs of LeRobot's recipes compute under bf16 autocast, which the documented commands leave
off; that changes how the fp32 weights (and SmolVLA's attention scores) compute, not how any weight is stored. We
compared the settings at the same learning rate. Because bf16 storage loses small updates, it acts like a lower
effective learning rate, which falls further as the learning rate falls (Section~\ref{sec:rounding}) and so also fits
the stall at $10^{-5}$. Whether bf16 storage at a higher learning rate matches the fp32 copy at its best rate is
untested for VLAs. In a 187M-parameter language model (Mamba-3) fine-tuned on GSM8K, an fp32 copy ended with a higher
held-out loss than bf16 storage at a learning rate of $2\times10^{-4}$ (five seeds) but a lower one at $10^{-4}$ and
$5\times10^{-5}$ (three seeds, exploratory) \cite{wheresmyupdate}. A study of RL fine-tuning traced its sparse weight
updates to bf16 training and reports identical performance with float32, without giving numbers \cite{a2509.04259}.

For practitioners: cast the trainable weights to fp32 before building the optimizer and compute in bf16 under autocast
(the box on page~1). In LeRobot, set \texttt{-{}-policy.\allowbreak dtype=float32} for $\pi_{0}$/$\pi_{0.5}$ even though
it is the default, because \texttt{-{}-policy.\allowbreak path} keeps the bf16 of a checkpoint saved with the flag; our
$\pi_{0.5}$ runs did not use the box's flags but cast the weights that the bf16 flag loads to fp32. On our RTX 5060 the
one-line SmolVLA change stores all 99.9M trainable weights in fp32, bit-identical to the cast in our runs; 200 training
steps under bf16 autocast then log the same losses as our fp32-copy setting, and it also trains without autocast, as
LeRobot's documented command runs. The copy costs under 1\,GB for SmolVLA (0.8\,GB more peak memory at batch 16); for
$\pi_{0.5}$ the fp32 weights and AdamW states fit on one 80\,GB GPU at batch 32 with gradient checkpointing when AdamW
updates one tensor at a time (about 75\,GB at peak), an option LeRobot's optimizer config does not expose. The copy
makes the effective step larger, so a learning rate tuned under bf16 storage may need lowering; in RL, two single-seed
reports saw training collapse at the unchanged rate \cite{a2609.28838,rlinf1649}. Where memory forbids an fp32 copy,
carry the rounding error in a bf16 residual, as LeRobot's MolmoAct2 optimizer does \cite{lerobot4249}, or round
stochastically \cite{a2010.06192}. We tested neither, but in fine-tuning of ternary language models, emulated versions
of both were non-inferior to fp32 training, in one run per setting \cite{a2610.07853}. Either way, compute $E(\lr)$
from the weights before training and run the probe during it.

\textbf{Availability.} The code, every training log, the outcome of every try, and the dated project log with every
rule, amendment and deviation are available from the author on request; trained checkpoints were not kept.

\IEEEtriggeratref{25}
\bibliographystyle{IEEEtran}
\bibliography{refs}

\end{document}